\documentclass[11pt]{article}
\usepackage[final]{coling}

\usepackage{times}
\usepackage{booktabs}
\usepackage{latexsym}
\usepackage{pdfpages}

\usepackage[utf8]{inputenc} 
\usepackage{microtype}
\usepackage{inconsolata} 
\usepackage{hyperref} 
\usepackage{fontawesome5}
\usepackage{colortbl}
\usepackage{xcolor}
\usepackage{amsmath} 
\usepackage{tikz}
\usepackage{pgfplots}
\pgfplotsset{compat=1.17}
\usepackage{pgf-pie}

\usepackage{xcolor}
\usepackage{svg}

\usepackage{graphicx}

\title{NLPineers@ NLU of Devanagari Script Languages 2025: Hate Speech Detection using Ensembling of BERT-based models }

\author{
    Anmol Guragain$^{1}$\thanks{Equal contribution.}, 
    Nadika Poudel$^{1}$\footnotemark[1], 
    Rajesh Piryani$^{2}$, 
    Bishesh Khanal$^{1}$ \\
    $^{1}$Nepal Applied Mathematics and Informatics Institute for Research (NAAMII), Nepal \\
    $^{2}$IRIT, Universite Toulouse III Paul Sabatier (UT3), Toulouse, France
}

\begin{document}
\maketitle

\begin{abstract}
This paper explores hate speech detection in Devanagari-scripted languages, focusing on Hindi and Nepali, for Subtask B of the CHIPSAL@COLING 2025 Shared Task. Using a range of transformer-based models such as XLM-RoBERTa, MURIL, and IndicBERT, we examined their effectiveness in navigating the nuanced boundary between hate speech and free expression. Our best performing model, implemented as ensemble of multilingual BERT models achieved Recall of 0.7762 (Rank 3/31 in terms of recall) and F1 score of 0.6914 (Rank 17/31). To address class imbalance, we used backtranslation for data augmentation, and cosine similarity to preserve label consistency after augmentation. This work emphasizes the need for hate speech detection in Devanagari-scripted languages and presents a foundation for further research. The code can be accessed at \href{https://github.com/Anmol2059/NLPineers}{https://github.com/Anmol2059/NLPineers}.
\end{abstract}

\section{Introduction}

Social media has become an essential part of our lives, empowering users to communicate freely and fostering a global exchange of ideas. However, it has contributed to the rapid proliferation of harmful content, including hate speech. Detecting hate speech is inherently complex due to the nuanced boundary between hate speech and legitimate free expression. What one individual may perceive as an offensive or harmful statement, another might interpret as a right to free speech, complicating the task of building an automated hate speech detection system. In languages where Devanagari script is predominantly used, such as Hindi, Marathi, and Nepali, detecting hate speech becomes even more intricate due to linguistic diversity, regional variations, and code-mixing practices.

Numerous transformer-based models have emerged to address the challenges of hate speech detection across various high-resource languages. HateBERT \citep{DBLP:journals/corr/abs-2010-12472}, a BERT model retrained on a dataset of Reddit comments from communities banned for offensive content, outperforms general BERT models in detecting abusive language in English. MC-BERT4HATE \citep{yang2020mcbert4hate} presents a multi-channel architecture that integrates English, Chinese, and multilingual versions of BERT, aiming to detect hate speech across multiple languages more effectively. However, during politically charged events like elections, hate speech in Devanagari scripts intensifies on social media platforms like Twitter, producing more complex forms that require an understanding of socio-political dynamics beyond mere linguistic processing. These nuances are not well captured by general-purpose models, highlighting the need for specialized approaches.

The First Workshop on South East Asian Language Processing \citep{sarves2025chipsal} aims to strengthen and spur NLP research and development in SEA languages. This paper aims to solve Task B: Hate Speech Detection of the Shared Task on Natural Language Understanding of Devanagari Script Languages \citep{thapa2025nludevanagari}. Hate speech detection is a binary classification problem that requires determining whether a tweet is hate speech or not. The classifiers we used in this challenge include XLM-RoBERTa, MURIL, and IndicBERT.

\section{Related Works}
Devanagari-script languages, being low-resource, have seen relatively limited work in hate speech detection. Aggression and Misogyny Detection using BERT by \citep{safi-samghabadi-etal-2020-aggression} classified comments presents in English, Hindi and Bengali into one of the three aggression classes - Not Aggressive, Covertly Aggressive, Overtly Aggressive, as well as one of the two misogyny classes - Gendered and Non-Gendered scoring 0.8579 weighted F1-measure using BERT model. In second workshop on Trolling, Aggression, and Cyberbullying (TRAC-2), \citep{baruah-etal-2020-aggression} work on Shared task on Misogynistic Aggression Identification achieved highest F1 score of  0.87 in Hindi language using XLMRoBERTa . Similarly, HASOC 2020: Hate Speech and Offensive Content Identification in Indo-European Languages \citep{mandl2020hasoc} had sub-task for Hate Speech detection in Hindi, German and English having 40 teams as participants. The best submission for Hindi used a CNN with fastText embeddings as input and the best result for English is based on a LSTM which used GloVe embeddings as input. Although there has been some work done on Hindi, it is worth noting that Nepali, which also uses the Devanagari script, has received relatively little attention in this area,\citep{luitel2024can,niraula2022linguistic} likely due to resource limitations. \citep{niraula-etal-2021-offensive} annotated 7462 records in Devanagari scripts into four categories SEXIST, RACIST, OTHER-OFFENSIVE, and NON-OFFENSIVE using Random Forest Classifier achieving F1 scores as 0.01, 0.45,0.71 and 0.87 respectively. 

Despite these few efforts, there remains a lack of performance benchmarks for multilingual BERT models on Devanagari scripts. Additionally, previous works have not explored BERT-based ensemble strategies that integrate predictions from multiple models. This gap motivated us to investigate the effectiveness of various multilingual BERT models and their ensembling approaches for hate speech detection in Devanagari scripted languages.

\section{Dataset and Task}

\begin{table}[h]
\centering
\small
\begin{tabular}{@{}lrr@{}}
\toprule
\textbf{Category} & \textbf{Training} & \textbf{Evaluation} \\
\midrule
Hindi Non-Hate & 7,376 & 1,596 \\
Nepali Non-Hate & 9,429 & 2,006 \\
Hindi Hate & 679 & 142 \\
Nepali Hate & 1,535 & 332 \\
\midrule
\textbf{Total} & 19,019 & 4,076 \\
\bottomrule
\end{tabular}
\caption{Sample distribution of data in training and evaluation sets}
\label{tab:dataset_distribution}
\end{table}

\begin{figure}[t]

    \centering
    \includegraphics[width=0.5\textwidth]{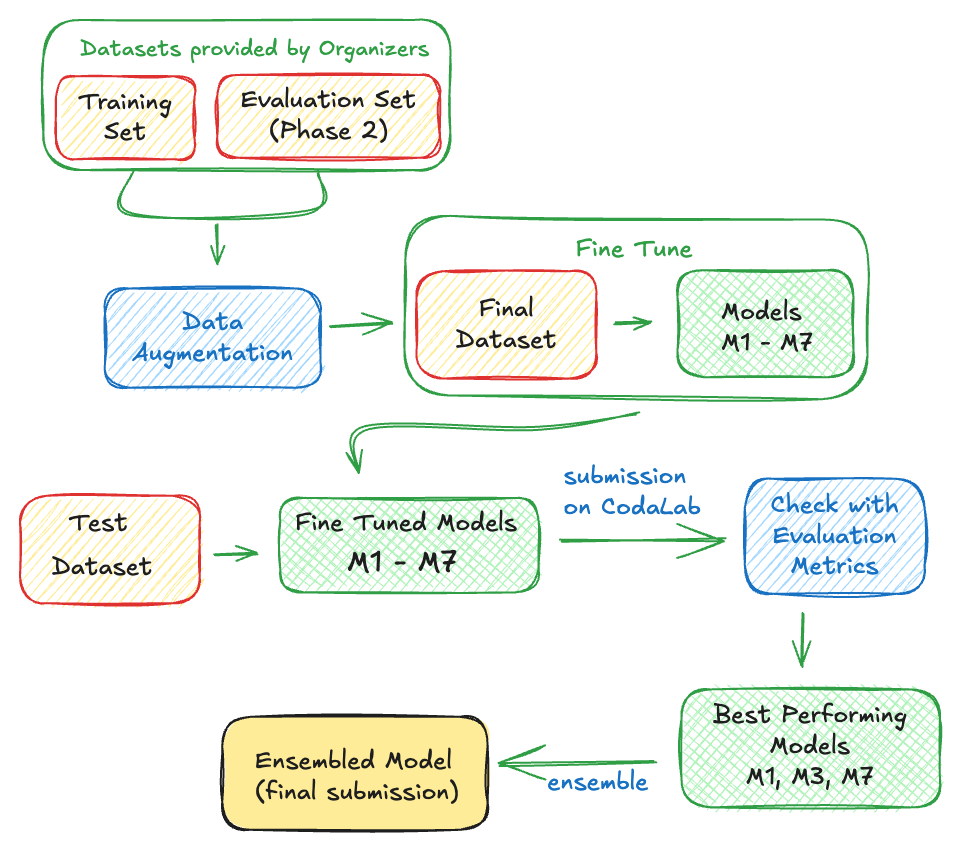}
    \centering
    \caption{Experiment Workflow.}
    \label{fig:flow_chipsal}
\end{figure}

This experiment uses the data from the shared task, which is compiled from prior works across multiple Devanagari-script languages, including Hindi hate speech in political discourse \citep{jafri2024chunav, jafri2023uncovering}, Nepali election discourse \citep{thapa2023nehate, rauniyar2023multi}, Bhojpuri-English sentiment modeling \citep{ojha2019english}, Marathi sentiment analysis \citep{kulkarni2021l3cubemahasent}, and Sanskrit translation corpora \citep{aralikatte2021itihasa}. In this study, the evaluation set refers to the phase 2 data provided by the challenge organizers, which is part of the development data. It is distinct from the test set, which was submitted for the challenge.  Table \ref{tab:dataset_distribution} provides detailed statistics on the original dataset of the shared task and some of the example sentences are in Figure \ref{fig:datasets_examples}. The overall pipeline of this experiment is summarized on Figure \ref{fig:flow_chipsal}.

\section{Experimental Setup}

\begin{table*}
\small
\centering
\begin{tabular}{@{}clll@{}}
\toprule
\textbf{} & \textbf{Model}                                & \textbf{Tokenizer / Embedding}             & \textbf{Classifier Architecture}               \\ \midrule
M1             & MuRIL abusive \citep{das2022data}                 & MuRIL (Hindi-Abusive) - Self               & Native Head     \\
M2             & MuRIL + TabNet              & MuRIL - Self                               & TabNet Classifier     \\
M3             & MuRIL\citep{khanujaMuril}& MuRIL - Self                               & Native Head                    \\
M4             & IndicBERT\citep{kakwani2020indicnlpsuite}                   & IndicBERT - Self                           & Native Head         \\
M5             & IndicBERT + LSTM-CNN        & IndicBERT - Self                           & LSTM + CNN + FC layer                                    \\
M6             & XLM-Roberta + Logistic Regression & XLM-Roberta - Self                         & Logistic Regression                            \\
M7             & XLM-Roberta \citep{conneau2020unsupervisedcrosslingualrepresentationlearning} & XLM-Roberta - Self                         & Native Head  \\
M8             & FastText + LSTM           & None - FastText (Hindi + Nepali)           & LSTM  + FC Layer                                         \\ \bottomrule
\end{tabular}%

\caption{Overview of models used. \textbf{FC} stands for Fully Connected layer, and \textbf{Native Head} refers to the model’s built-in classification head when imported from Hugging Face, indicating that these models are fine-tuned. The \textbf{Tokenizer/Embedding} column combines the tokenization method and embedding source; “Self” signifies that embeddings are generated by the model itself. Tokenizers and models are sourced from Hugging Face’s model hub, with FastText embeddings from FastText.cc.
}
\label{tab:model_tokenizer_classification_head}
\end{table*}

\subsection{Data Augmentation}

As observed in Table~\ref{tab:dataset_distribution}, hate speech instances were much fewer compared to non-hate speech. To address this imbalance, data augmentation was applied to the hate speech instances using backtranslation with the  mBART-large-50 model \citep{DBLP:journals/corr/abs-2008-00401}, translating the data to English and back to the source language to introduce text variations. To ensure that the augmented data retained semantic similarity with the original data and minimize risk of unintended label changes, we calculated the cosine similarity between the embeddings of the augmented and original texts using the XLM-RoBERTa base model \citep{DBLP:journals/corr/abs-1911-02116}. Only augmented data with a cosine similarity score greater than 0.9(chosen empirically) was added to the final dataset.

For the training set, data augmentation was performed on both Hindi and Nepali hate speech instances. In the evaluation set, however, augmentation was applied only to Hindi instances, as Nepali data already had a higher representation compared to Hindi data. Additionally, to further address the class imbalance, all hate speech instances (label '1') were duplicated to give the model more exposure to minority class. After augmentation, the training set grew to 13,695 instances by incorporating the original 2,214 training, 474 evaluation, and their augmented instances.

\subsection{Pretrained Models}
We used different pre-trained models and classifier heads, as observed in Table~\ref{tab:model_tokenizer_classification_head} to explore approaches that could better capture the nuances that exist in recognizing of hate speeches. The models included three BERT-based architectures—MuRIL \citep{DBLP:journals/corr/abs-2103-10730}, XLM-RoBERTa \citep{conneau2020unsupervisedcrosslingualrepresentationlearning}, and IndicBERT \citep{kakwani2020indicnlpsuite}—as well as FastText \citep{grave2018learning}, an token embedding-based model. MuRIL is pre-trained on 17 Indian languages and their transliterations. XLM-RoBERTa, a large multilingual model, offers cross-lingual capabilities by training on diverse data from multiple languages. IndicBERT focuses on 12 Indian languages, including Devanagari(Hindi and Marathi), and uses a lightweight structure ideal for efficient processing. In contrast, FastText uses character-level n-grams to provide a detailed lexical representation, which is particularly beneficial for morphologically rich languages in Devanagari script. This combination allows us to leverage both deep contextual understanding and fine-grained lexical details for effective hate speech detection.

\subsection{Ensemble Strategy}

Our ensemble strategy leveraged the strengths of our top-performing models from Table \ref{tab:results}. We chose M7 (XLM-Roberta) as the primary model, M3 (MuRIL) as the secondary model, and M1 (MuRIL abusive) as the fallback model, based on each model’s unique strengths.

\begin{itemize}
    \item \textbf{Primary Model (XLM-Roberta, Model 7):} XLM-Roberta achieved the highest recall (0.7381), making it effective at detecting hate speech and minimizing missed cases. 
    
    \item \textbf{Secondary Model (MuRIL, Model 3):} When XLM-Roberta does not predict hate speech, MuRIL provides a balanced F1 score (0.6904) and accuracy (0.8744), acting as a secondary layer to catch potential cases missed by the primary model.
    
    \item \textbf{Fallback Model (MuRIL abusive, Model 1):} In cases where both primary and secondary models predict no hate speech, MuRIL abusive, with the highest precision (0.7572) and accuracy (0.8950), serves as a conservative fallback to minimize false positives.
   
\end{itemize}
\small
\[
\text{prediction}(x) = 
\begin{cases} 
    1 & \text{if } \text{M}_7(x) = 1 \\
    1 & \text{if } \text{M}_7(x) = 0 \text{ and } \text{M}_3(x) = 1 \\
    \text{M}_1(x) & \text{otherwise}
\end{cases}
\]
\normalsize

\begin{table*}[ht]
    \centering
    
    \vspace{0.1 in}
    
    \begin{tabular}{llcccc}
        \toprule
        \textbf{} & \textbf{Model} & \textbf{Recall} & \textbf{Precision} & \textbf{F1 Score} & \textbf{Accuracy} \\
        \midrule
        M1 & MuRIL abusive & 0.6335 & \cellcolor{green!40}0.7572 & 0.6681 & \cellcolor{green!40}0.8950 \\
        M2 & MuRIL + TabNet & 0.6296 & 0.5874 & 0.5984 & 0.7927 \\
        M3 & MuRIL & \cellcolor{green!10}0.6877 & 0.6934 & \cellcolor{green!10}0.6904 & \cellcolor{green!20}0.8744 \\
        M4 & IndicBERT & 0.5934 & 0.5915 & 0.5924 & 0.8305 \\
        M5 & IndicBERT + LSTM-CNN & 0.6455 & 0.5618 & \cellcolor{red!40}0.5207 & \cellcolor{red!40}0.6271 \\
        M6 & XLM-Roberta + Logistic Regression & 0.6504 & 0.6596 & 0.6548 & \cellcolor{green!10}0.8619 \\
        M7 & XLM-Roberta  & \cellcolor{green!40}0.7381 & \cellcolor{green!20}0.6696 & \cellcolor{green!40}0.6933 & 0.8472 \\
        M8 & FastText + LSTM & \cellcolor{red!40}0.5320 & \cellcolor{red!40}0.5400 & 0.5346 & 0.8270 \\
        \midrule
         & Ensemble (M1, M3, M7) &\cellcolor{green!80}0.7762 & \cellcolor{green!10}0.6639 & \cellcolor{green!20}0.6914 & 0.8258 \\
        \bottomrule
        
    \end{tabular}
    \caption{Evaluation results on test set of the hate speech detection task. Dark green cells indicate the best performance in the respective metric, while dark red cells indicate the worst. Gradual shades of green represents relatively good performance.}
    \label{tab:results}
    
\end{table*}

\subsection{Hyperparameters and Compute Environment}
Training utilized the following hyperparameters, determined through iterative testing and practical constraints: a learning rate of 2e-5, a batch size of 16. These values were selected to balance model performance with available compute resources and processing time. 
We used NVIDIA GeForce RTX 3090 as compute environment.


\section{Results and Discussions}

The competition was hosted on the Codalab\footnote{\href{https://codalab.lisn.upsaclay.fr/competitions/20000\#participate-submit_results}{https://codalab.lisn.upsaclay.fr/competitions/20000\#participate-submit\_results}} platform by the organizers, where we submitted binary predictions (0 or 1) for evaluation based on recall, precision, F1 score, and accuracy. The performance of our models in the test set of challenge is shown in Table~\ref{tab:results}.

\subsection{MuRIL-Based Models}  
As seen in Table~\ref{tab:results}, the fine-tuned MuRIL model (M1) on Devanagari script provided the highest accuracy among MuRIL-based models. The standard MuRIL model (M3) demonstrated balanced performance across all metrics. We also experimented with combining MuRIL and TabNet (M2) as suggested by \citep{chopra2023framework}, but this configuration did not yield competitive results in this task.

\subsection{IndicBERT}  
We anticipated strong performance from IndicBERT (M4) due to its training on 12 Indic languages as discussed in \citep{kakwani2020indicnlpsuite}. However, its results were lower than expected, possibly due to the presence of Nepali text  in the dataset, which IndicBERT may not be optimized for. The combination of IndicBERT with LSTM-CNN also underperformed, showing unsatisfactory results.

\subsection{XLM-Roberta-Based Models}  
Both the plain XLM-Roberta model (M7) and XLM-Roberta with a Logistic Regression head (M6) performed well, indicating the model’s robust generalization capabilities across different metrics. This highlights XLM-Roberta’s versatility in multilingual tasks.

\subsection{FastText with LSTM}  
Since the evaluation set contained Devanagari script for both Nepali and Hindi, we utilized FastText embeddings of both languages and fed in LSTM based classifer  (M8). However, this setup did not yield satisfactory results, likely due to the limitations of static embeddings, which struggle to capture the contextual nuances essential for accurate hate speech detection.

\subsection{Ensembled Model}  
Our final submission was an ensemble model combining M1, M3, and M7, as described in previous sections. This ensemble achieved balanced performance, with recall, precision, F1 score, and accuracy of 0.7762, 0.6639, 0.6914, and 0.8258, respectively, effectively leveraging the strengths of multiple models.

\section{Conclusion}
This study highlights the potential of  various BERT-based models and ensembling approach for hate speech detection in Devanagari-scripted languages, with future work planned on model robustness and scalability for real-world applications. Further research could explore additional embeddings and augmentations to enhance performance across multilingual contexts.
\clearpage
\section*{Limitations}

This study faces several limitations, particularly due to the linguistic complexities inherent to Devanagari-scripted languages like Hindi and Nepali. Below, we outline some of the primary challenges:

\begin{itemize}

    \item \textbf{Limitations of Data Augmentation via Backtranslation}: 
    While backtranslation with the mBART model was used to augment hate speech samples, this approach sometimes loses the cultural nuances or tone intended in the original text. For instance, words like \textbf{\textit{tapai}} and \textbf{\textit{hajur}} in Nepali convey a formal or respectful tone, but during translation to English and back to Nepali, these terms are often reduced to the informal \textbf{\textit{timi}}, altering the sentiment. This limitation could introduce subtle inaccuracies during model training.

    \item \textbf{Contextual Meaning Across Languages}: In Devanagari-scripted languages, certain words can carry vastly different meanings depending on the language context. Such linguistic ambiguities create challenges for the model, as it may misinterpret hate speech in cases where meanings differ across languages using the same script.


    \item \textbf{Dependency on Word Embeddings for Devanagari Script}: 
    Devanagari script is used for multiple languages, and words in Hindi and Nepali may have similar or identical representations in embeddings, potentially leading to confusion. While BERT-based models like XLM-RoBERTa and MuRIL are designed to handle multilingual contexts, challenges persist when languages share the same script but differ in vocabulary or syntax. These issues may impact the model's ability to differentiate nuanced expressions unique to each language.

\end{itemize}

\section*{Acknowledgements}
We thank CHIPSAL@COLING organizers for providing annotated datasets and hosting the shared task. We also thank NepAl Applied Mathematics and Informatics Institute for research(NAAMII) for giving us access to training resources required for the sub-task.

\appendix

\appendix

\begin{figure*}[htbp]
\section{Appendix}
\subsection{Example Sentences from Dataset}

    \centering
    \includegraphics[width=\textwidth]{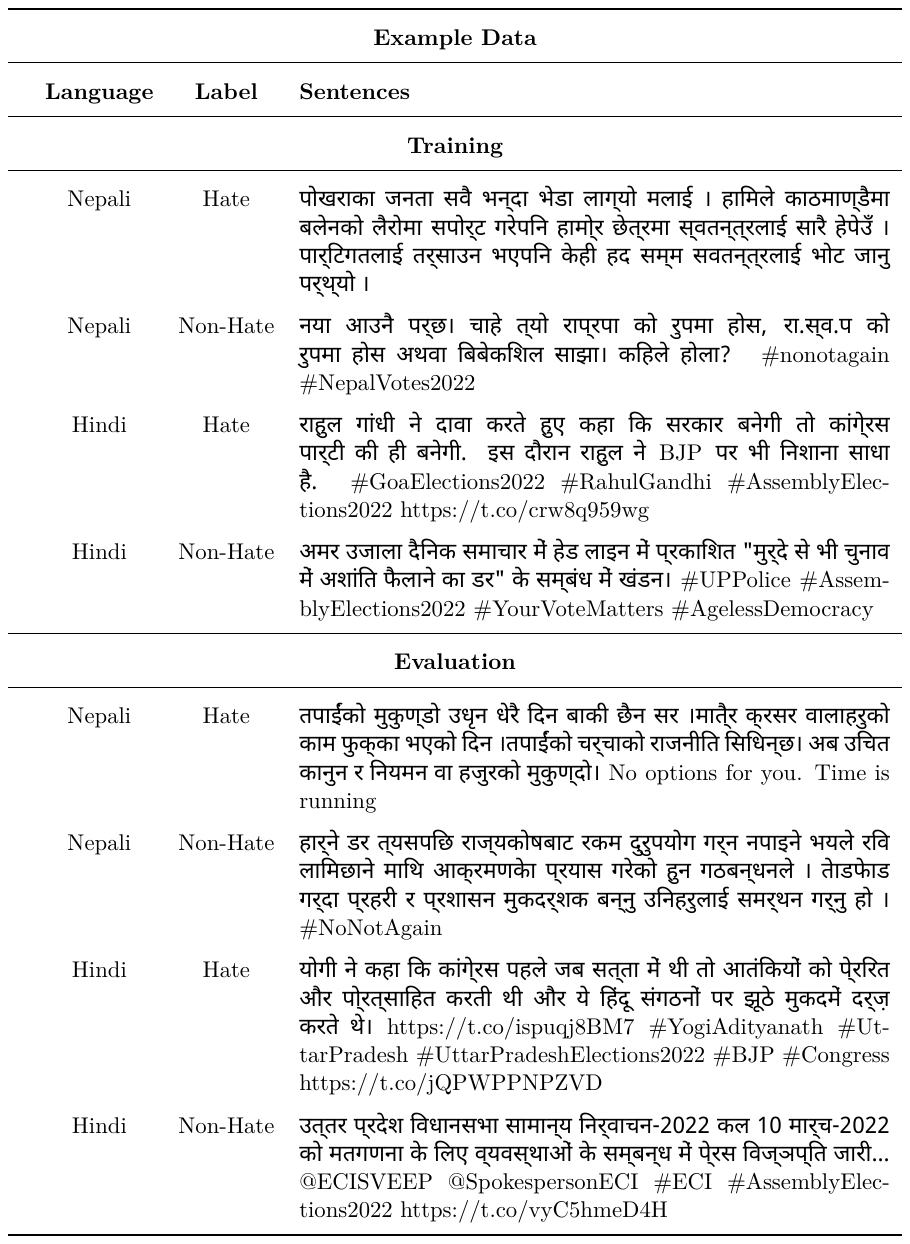} 
    \caption{Examples from the dataset.}
    \label{fig:datasets_examples}
\end{figure*}




\end{document}